\documentclass{article}

\usepackage{PRIMEarxiv}

\usepackage[utf8]{inputenc} % allow utf-8 input
\usepackage[T1]{fontenc}    % use 8-bit T1 fonts
\usepackage{hyperref}       % hyperlinks
\usepackage{url}            % simple URL typesetting
\usepackage{booktabs}       % professional-quality tables
\usepackage{amsfonts}       % blackboard math symbols
\usepackage{nicefrac}       % compact symbols for 1/2, etc.
\usepackage{microtype}      % microtypography
\usepackage{lipsum}
\usepackage{fancyhdr}       % header
\usepackage{graphicx}       % graphics
\graphicspath{{media/}}     % organize your images and other figures under media/ folder
\usepackage{algorithm}% <=================http://ctan.org/pkg/algorithms
\usepackage{algpseudocode}% <========== http://ctan.org/pkg/algorithmicx
\usepackage{subfig}

\title{FLAG: Fast Label-Adaptive Aggregation for Multi-label Classification in Federated Learning
}

\author{
  Shih-Fang, Chang \\
  Information and Communications Research Laboratories\\
  Industrial Technology Research Institute \\
  Hsinchu\\
    \And
  Benny Wei-Yun Hsu \\
  Institute of Computer Science and Engineering \\
  National Yang Ming Chiao Tung University \\
  Hsinchu\\
    \And
  Tien-Yu Chang \\
  Institute of Computer Science and Engineering \\
  National Yang Ming Chiao Tung University \\
  Hsinchu\\
   \AND
   Vincent S. Tseng* \\
   Department of Computer Science \\
   National Yang Ming Chiao Tung University, Hsinchu, Taiwan, R.O.C \\
   orcid=0000-0002-4853-1594 \\
   \texttt{vtseng@cs.nycu.edu.tw} \\
}

\begin{document}
\maketitle

\begin{abstract}
Federated learning aims to share private data to maximize the data utility without privacy leakage. Previous federated learning research mainly focuses on multi-class classification problems. However, multi-label classification is a crucial research problem close to real-world data properties. Nevertheless, a limited number of federated learning studies explore this research problem. Existing studies of multi-label federated learning did not consider the characteristics of multi-label data, i.e., they used the concept of multi-class classification to verify their methods' performance, which means it will not be feasible to apply their methods to real-world applications. Therefore, this study proposed a new multi-label federated learning framework with a Clustering-based Multi-label Data Allocation (CMDA) and a novel aggregation method, Fast Label-Adaptive Aggregation (FLAG), for multi-label classification in the federated learning environment. The experimental results demonstrate that our methods only need less than 50\% of training epochs and communication rounds to surpass the performance of state-of-the-art federated learning methods.
\end{abstract}

% keywords can be removed

\section{Introduction}\label{1}
Recently, privacy and data security have been two crucial topics in deep learning applications. Applying deep learning techniques to industries must consider protecting users' private information (e.g., personal health data). Federated learning has been proposed to take advantage of private data to maximize the data utility without data leakage. By training on local devices (clients) and collaboratively learning a whole federated model through aggregate local knowledge, federated learning is able to train the network without direct access to private local data. With the federated learning methods, clients can achieve higher performance and faster convergence speed than training on their own local datasets.

FedAvg \cite{FedAvg} is the most well-known federated learning method, which leads federated learning to be more practical by effectively reducing communication rounds. FedAvg trains client models on client local data and aggregates them by averaging their parameters. Then, Fedprox \cite{Fedprox} first considers system heterogeneity and client data heterogeneity in federated learning. After that, many studies were proposed to improve the client performance and converge speed under client data heterogeneity in federated leaning \cite{surveyPFL}. For instance, Per-FedAvg \cite{PerFedAvg} and pfedHN \cite{pFedHN} introduce meta-learning into federated learning framework. Some studies focus on utilizing personalized layers to overcome clients data heterogeneity in federated learning \cite{siloBN} \cite{MTFL} \cite{fedbn} \cite{fedpc}. Other studies including PFADET \cite{PFADET}, FedDF \cite{NEURIPS2020_feddf}, and KTpFL \cite{KTpFL} focused on improving the procedure of aggregating local models through knowledge distillation. One of the most critical challenges of federated learning is the
heterogeneity of the client data, as data distributed Non-Independent and identically distributed (non-i.i.d.) across clients. A recent study \cite{surveyhetro} divides client heterogeneity into three major types: quantity skew, label distribution skew, and acquisition skew. This research concentrates on the label distribution skew problem in federated learning.

Former federated learning research usually considered multi-class classification for applications. However, multi-label classification is the other crucial research problem. In the meantime, multi-label data are much closer to real-world features. Existing multi-label classification works involve many unique research problems, such as label correlation, label error correction, and data imbalance. Also, multi-label classification techniques can be used in several practical applications—for example, object detection and multi-label disease detection. There are two common approaches to solving the above problems: cost-sensitive approaches and re-sampling  \cite{review_ml} \cite{dbloss} \cite{ASL}. Nevertheless, these methods may not be appropriate for modeling under a federated mechanism. In this study, we devoted ourselves to developing novel methods for multi-label classification in the federated learning environment.

Although multi-label federated learning is a critical research problem and valuable in real-world applications, only a few studies consider multi-label classification in federated learning. Two major challenges exist in developing a multi-label federated learning method: (1) design of experiments (DOE) for multi-label classification in federated learning; (2) heterogeneity of multi-label distribution in clients. Conventional multi-class federated learning used Dirichlet distribution to simulate client heterogeneity. However, it cannot be applied to multi-label federated learning because each subject has greater than or equal to one label, leading to different distributions from multi-class data. This study aims to develop a comprehensive simulation for the heterogeneity of clients and consider multi-label characteristics in federated learning, including imbalance of label distribution and frequency.

The contributions of this research can be summarized as follow:
\begin{itemize}
\item{} We proposed a novel method, FLAG (Fast Label-adaptive AGgregation), for multi-label federated learning. FLAG considered multi-label distribution and correlation of each client to aggregate high-performance models.
\item{} We presented a Clustering-based Multi-label Data Allocation (CMDA) method that is the first to consider simulating multi-label data distribution of clients in federated learning.
\item{} Our proposed method outperformed the state-of-the-art federated learning methods with mean average precision (mAP) on the multi-label dataset. Meanwhile, it can save more than 50\% training epochs and communication rounds, while our method achieved greater than or equal to the mAP of the other methods.
\end{itemize}

The rest of this paper is organized as below. Section \ref{sec2} introduces the related work of personalized federated learning and multi-label learning. Section \ref{sec3} presents the proposed methods, including a new simulation method of experimental setups and a novel label-weighted aggregation method for multi-label federated learning. Later in Section \ref{sec4} and \ref{sec5}, we show a series of experiments to evaluate our proposed method and the state-of-the-art (SOTA) federated learning methods on the multi-label image dataset, MS-COCO \cite{MSCOCO}, demonstrating the effectiveness and efficiency of our method. Finally, in Section \ref{sec6}, we discuss and summarize this study's contribution and limitations.

\section{Related Work}\label{sec2}

\subsection{Personalized Federated learning}
Federated learning was first proposed by \cite{konevcny2016federated}, where they introduced structured and sketched updates to reduce the communication cost of federated learning. Then, \cite{FedAvg} proposed a simple and the most well-known framework FedAvg, which updates client models by simply averaging over clients' model weights and replacing them. FedAvg has been the most common method in federated learning and has successfully applied to multiple types of data, including image, time-series, or tabular datasets \cite{reviewFLapp} \cite{FLapp}. After that, \cite{Fedprox} pointed out two main shortages about FedAvg in real-world applications. Including system heterogeneity and client data heterogeneity, which vary the computation power and the data distribution of clients, respectively. Further, they proposed Fedprox, which introduced dynamic epochs and proximal term regulation to overcome the above problems. 

Recently, improving the performance of client models under data heterogeneity has become important research in federated learning. Per-FedAvg \cite{PerFedAvg} was the first to introduce meta-learning into the federated learning framework for training personalized client models under client data heterogeneity to avoid performance loss from client data heterogeneity. pfedHN incorporated Hypernetwork \cite{pFedHN} into the federated learning framework to learn the initialized parameters globally for each client and train personalized client models locally. However, they may only work in larger models if there exists a huge search space. Some studies focused on utilizing personalized layers to overcome clients' data heterogeneity in federated learning, such as \cite{Fedplayer} and \cite{fedpc} using penalization classifier; \cite{MTFL}, \cite{fedbn}, and \cite{siloBN} used personalized batch normalization techniques.
Additionally, PartialFed-Adaptive \cite{PartialFed} conducted a detailed study about personalized and shared layers, and they proposed an adaptive method to load layers for each client. These personalized federated learning studies focused on domain personalized but were limited in the global model.

Other studies focus on aggregating without simply averaging over clients' model weights. \cite{PFADET} proposed PFADET that introduced Progressive Fourier Aggregation to averaging weights in the frequency domain and Deputy-Enhanced Transfer for the knowledge distillation process. Also, some studies utilized zero-shot learning \cite{Fedzeroshot_2021_CVPR}, and data augmentation techniques \cite{fed_da} \cite{fedmix} \cite{mihetro_da} for knowledge distillation.Moreover, \cite{KTpFL} fully applied the knowledge distillation concept to update client models by learning clients' soft prediction on public data and the correlation between clients to overcome client data heterogeneity. However, These methods require additional datasets, a vast space for the models, and may not be attainable for more complicated datasets (e.g., multi-label data).

Over the years, there have been plenty of studies in the federated learning field. However, most of them only evaluated their methods on standard multi-class datasets; there are a limited number of studies regarding client data heterogeneity in multi-label data for federated learning.

\subsection{Multi-label classification in federated learning}
Previous studies have developed multi-label learning methods to solve multi-label classification problems and explore correlations between labels. There are two popular approaches to solving multi-label problems: 1) cost-sensitive methods and 2) resampling methods. Cost-sensitive methods use different cost metrics to describe the costs of different samples, aiming to balance multi-label datasets by minimizing the loss \cite{review_ml} \cite{ml_imbalance} \cite{dbloss}. Resampling methods are based on undersampling and oversampling \cite{review_ml} techniques. The resampling methods can be grouped by the sample selection approach \cite{review_ml} \cite{ml_resample} \cite{ml_resample2}. Recently, \cite{ASL} specified the most common label positive/negative problems and proposed Asymmetric Loss (ASL), which includes asymmetric focusing and asymmetric probability shifting
to overcome label imbalance and label error problems in multi-label learning. \cite{ml_transformer} aimed to discover label correlation and improve model ability by modifying the model; the common approaches are Conditional Prediction, Shared Embedding Space, Structured Output, and Label Graph.

Although federated learning and multi-label learning are both valuable topics in machine learning research, only a few works consider multi-label classification in federated learning. Moreover, most of them used multi-label datasets and evaluated their methods without considering multi-label data problems. \cite{mlfed} \cite{mlfed2} \cite{mlfed3}
To the best of our knowledge, only one study considers multi-label characteristics in federated learning \cite{mlfedgcn}. However, they took label correlation locally in each client without utilizing the advantage of federation learning and only compared their model with a limited number of methods. Besides, they did not mention how to divide the dataset into multi-clients. This paper mainly focuses on studying multi-label distribution problems in federated learning and presents a feasible simulation method for multi-label federated learning experiment settings.

\section{Methodology}\label{sec3}
This section introduces our proposed clustering-based multi-label data allocation (CMDA) for the client simulation in experiments and Fast Label-Adaptive Aggregation (FLAG) methods for multi-label federated learning. First, we present the concept and the algorithm regarding how to allocate the multi-label data to clients for a simulation of a federated learning environment with multi-label heterogeneity, and then we present the concept and detailed algorithm of FLAG.

\subsection{Proposed Framework}
\begin{figure*}[htbp]
	\centering
		\includegraphics[width=160mm,scale=1.0]{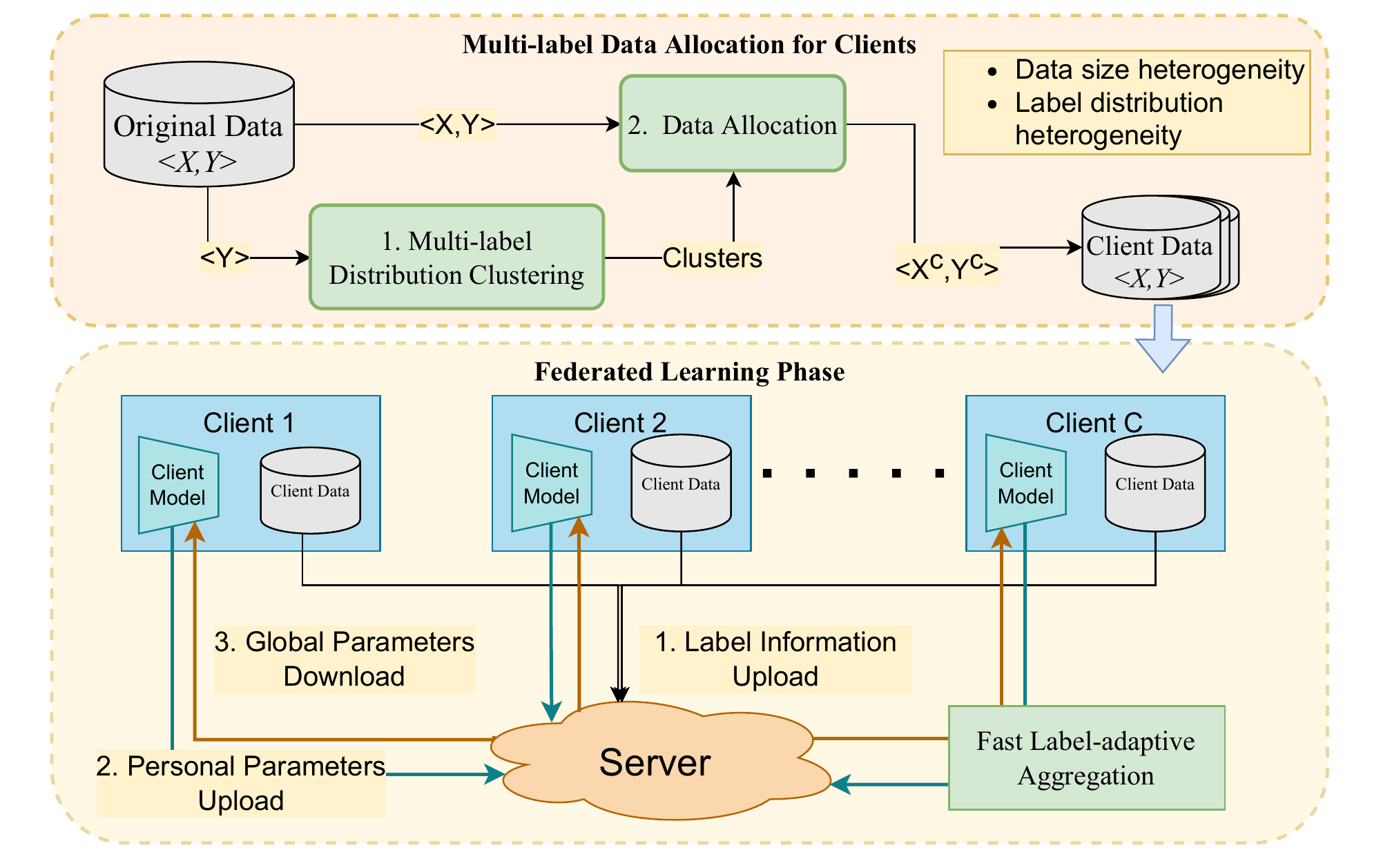}
	  \caption{The multi-label federated learning framework}\label{fig_framework}
\end{figure*}

Figure \ref{fig_framework} depicts the proposed multi-label federated learning framework. At first, the CMDA method uses the label distribution from original multi-label data as the input data for clustering. According to the input distribution, the clustering model generates the corresponding number of clusters for clients. For example, we aim to have C clients, so the number of clusters should be set to C. Therefore, we can allocate the clusters to clients to simulate a multi-label data distribution in the federated learning environment.

In the federated learning phase, 1) each client updates label statistics to the server without privacy or sensitive information. 2) each client uses their own data for local training, and then these clients upload model parameters to the server. At this stage, FLAG aggregates the uploaded label information and client model parameters for the global model update. 3) clients download the global model parameters from the server to update the local models. After several communication rounds and the training loss convergent, we evaluate client and global models at the end of training epochs and the aggregation stage.

\subsection{Clustering-based Multi-label Data Allocation}
In previous federated learning research, one of the crucial challenges was client heterogeneity. However, most datasets in the deep learning community are centralized datasets that are difficult to simulate client heterogeneity. There are two well-known methods to divide a centralized dataset into multiple client datasets with client heterogeneity. One assigns specific classes for each client, and the other uses Dirichlet distribution to allocate different class samples to each client. They are both commonly accepted in the federated learning setting \cite{dirichlet_used}. However, both methods assume each class is independent, which conflicts with multi-label properties.
Hence, we proposed CMDA for the client data splitting for further construction of a multi-label federated learning framework, which makes different clients have different label distribution trends. For instance, in the medical field, patients in different hospitals or medical centers may have some common symptoms and diseases. Still, they would have other specific diseases due to the property of regions (e.g., industrial compositions).

The proposed CMDA method considers label correlation while clustering samples through their label sets and assigning samples to each client according to the clustering results. Compared to the previous federated learning simulation method for the multi-class dataset, our approach uses the clustering algorithm to find the label correlation and form the natural clusters from the multi-label dataset. Then, different label distributions are divided from the centralized dataset via clustering to simulate client heterogeneity. 

The steps of our clustering split are as shown below:
First, the labels of each data sample are represented in binary format $y_i \in \{0,1\}^L$ which $1$ is positive and $L$ is the total label number. 
Second, we take advantage of the labels from each training sample as the feature for the clustering algorithm and assign the number of clusters as our simulated client number. After that, we perform a clustering algorithm to divide training and validation sets into clusters. We use the k-modes algorithm as our clustering method in this work \cite{kmodes}. K-modes is an algorithm similar to k-means clustering but more suitable for categorical features. The process of k-modes is described as follows.

\begin{itemize}
\item{Step1:} It randomly initializes k center points, and then assigns samples to different clusters by minimizing its dissimilarity measure to the center.
For two sample X and Y with m categorical features, the dissimilarity measure is $$d(X,Y)=\sum_{j=1}^{m}\delta(x_j,y_j)$$
where
\begin{equation}
\delta(x_j,y_j)=
%\begin{cases}
\{1, x_j = y_j ; 0, x_j \neq y_j\}
%\end{cases}\\
\end{equation}
\item{Step2:} Update every center point's features by cluster mode values.
\item{Step3:} Repeat the previous steps to reallocate samples until all samples' dissimilarity measure converge.
\item{Step4:} Output the sample's clustering results and the dissimilarity measure.
\end{itemize}

After assigning cluster indices for all training data, we use the same center points for clustering validation data.
We build the client's training and validation dataset by the clustering results. We illustrate the algorithm with pseudo-code in algorithm 1.

\begin{algorithm}
  \begin{algorithmic}[1]
\State $<X\textsubscript{tr},y\textsubscript{tr}>$ is feature and label from training dataset $D\textsubscript{tr}$, and 
$<X\textsubscript{val},y\textsubscript{val}>$ is feature and label from validation dataset $D\textsubscript{val}$%\;
\State $y\textsubscript{tr}$ and $y\textsubscript{val}$ are labels in binary forms%\;  % <===========================
\State $N$ is the client number user set%\;
\State $Cluster_{tr}$ and $Cluster_{val}$ are the cluster index of training and validation datasets%\;
\State $Center_{tr}$ is the center of each cluster%\;
\State $D^c_{tr}$ and $D^c_{val}$ are output training and validation datset for client c%\;
\Procedure {Clustering-based Simulation} {$D\textsubscript{tr}$, $D\textsubscript{val}$, $N$}%\;
\State $Center_{tr} \gets kmodes.fit(D\textsubscript{tr},N)$
\State $Cluster_{tr} \gets kmodes.transform(D\textsubscript{tr},Center_{tr},N)$
\State $Cluster_{val} \gets kmodes.transform(D\textsubscript{val},Center_{tr},N)$
  \For {$c\gets 1 \to N$} % <=======================
    \State $D^c_{tr} \gets \{\}$
    \State $D^c_{val} \gets \{\}$
    \For {$i \gets 1 \to |D_{tr}| $}
      \State $<X^i,y^i> \gets D_{tr}^{i}$
      \If{$Cluster_{tr}^i = c$}
      \State $Client^c_{tr} \gets Cluster_{tr} \cup \{<X^i,y^i>\}$
      \EndIf
    \EndFor
    \For {$i \gets 1 \to |D_{val}| $}
      \State $<X^i,y^i> \gets D_{val}^{i}$
      \If{$Cluster_{val}^i = c$}
      \State $Client^c_{val} \gets Cluster_{val} \cup \{<X^i,y^i>\}$
      \EndIf
    \EndFor
  \EndFor
  \State return $\{Client^c_{tr} , 1 \leq c \leq N \}$, $\{Client^c_{val} , 1 \leq c \leq N \}$
\EndProcedure
\caption{Clustering-based Client Simulation}\label{cluster_fedsimulate}
\end{algorithmic}
\end{algorithm}

\subsection{Fast Label-Adaptive Aggregation}
In a multi-label dataset, each sample contains multiple labels, and these labels usually do not occur independently and uniformly, i.e., there exists some correlation, specific frequency, and distribution among labels. It implies that features and labels possess critical information in a multi-label dataset that can enhance learning effectiveness. To capture this crucial property and protect privacy in federated learning, we proposed a label distribution weighted aggregation method, named Fast Label-adaptive AGgregation (FLAG), that weights client aggregation via label distribution and occurrence.

At the beginning of the communication round, each client calculates their label weights locally. The label weight calculation is based on 1) label distribution and 2) label occurrence in the client dataset. Label distribution is how many positive labels are present over all possible labels, and label occurrence defines the frequency of positive labels.
Under the proposed framework, clients locally calculate their label weights to avoid data leakage in federated learning. The definition of our label weight is as follows:
\begin{equation}
label\_weight = \{ \sum_{l=1}^{N_l}(\sum_{i=1}^{N_i^c} y_i^c)^\alpha , 1 \leq c \leq N_c\}
\end{equation}
Which $N_c$, $N_l$, and $N_i^c$ are the number of total clients, labels, and sample number of client c.
$\alpha$ is a hyper-parameter that controls the importance between label occurrence and distribution.
$\alpha = 0$ means label weight only considers label distribution wideness.
Moreover, the higher the $\alpha$ is, the more critical for label occurrence.
In our experiment, we set $0 \leq \alpha \leq 1$ to explore the influence of the parameters.

\section{Experimental Setup}\label{sec4}
This section describes the dataset, data allocation for clients, and evaluation metrics for experiments at first. Next, we introduce the backbone model for our method, baseline, and state-of-the-art methods for comparison. Finally, we demonstrate the results of the client data from the clustering-based multi-label data allocation method with k-modes to illustrate the client heterogeneity in the simulated federated environment. The experiment environment is implemented in Python3.8 and Pytorch 1.11.0. The device we use for all experiments is
Intel(R) Xeon(R) Gold 6154 CPU and Tesla V100-SXM2 GPU.

\subsection{Data Description and Evaluation Metrics}
We conducted a series of experiments with the MS-COCO 2014 multi-label dataset for our training and validation sets \cite{MSCOCO}. MS-COCO contains image classification, object localization, semantic segmentation, and individual object segmenting tasks. In this work, we only consider multi-label image classification tasks. We use the k-modes clustering method for the simulation of the client heterogeneity (i.e., the heterogeneity of multi-label data between clients) in federated learning. In the implementation, we first use the scikit-learn package to cluster training labels. After, divide both training and validation datasets into ten clients through the clustering results. The comparison between our CMDA method and random splitting shows in \ref{kmodes}. 
We took mean Average Precision (mAP) and convergent speed as the primary evaluation metrics for our experiment to verify the performance. We averaged all clients' mAP, demonstrated the worst client's mAP, and averaged global model mAP on clients' validation set to evaluate our method.

\subsection{Baseline and State-of-The-Art Models}
We followed the experiment settings of \cite{ASL} but in a federated environment. We took TRresNet \cite{tresnet} as our backbone model and Asymmetric Loss as the loss function. Adam optimizer and OneCycleLR scheduler were used for training.
The baseline we selected are Local client training TRresNet (TRresNet (L)) and FedAvg \cite{FedAvg}. The centralized learning (global model) TRresNet (TRresNet (G)) is the upper bound baseline.
In this study, the state-of-the-art (SOTA) methods included meta-learning-based methods, Per-FedAvg \cite{PerFedAvg} and pFedHN \cite{pFedHN}; personalized-layer-based methods, Personal batch normalization layer(Personal BN) and Personal classifier;
knowledge distillation methods, PFADET \cite{PFADET} and KT-pFL \cite{KTpFL}. All the above methods are representative federated learning methods. 

\subsection{k-modes Clustering-based Client Simulation}\label{kmodes}
\begin{figure}[htbp]
	\centering
	        \includegraphics[width=120mm,scale=1.0]{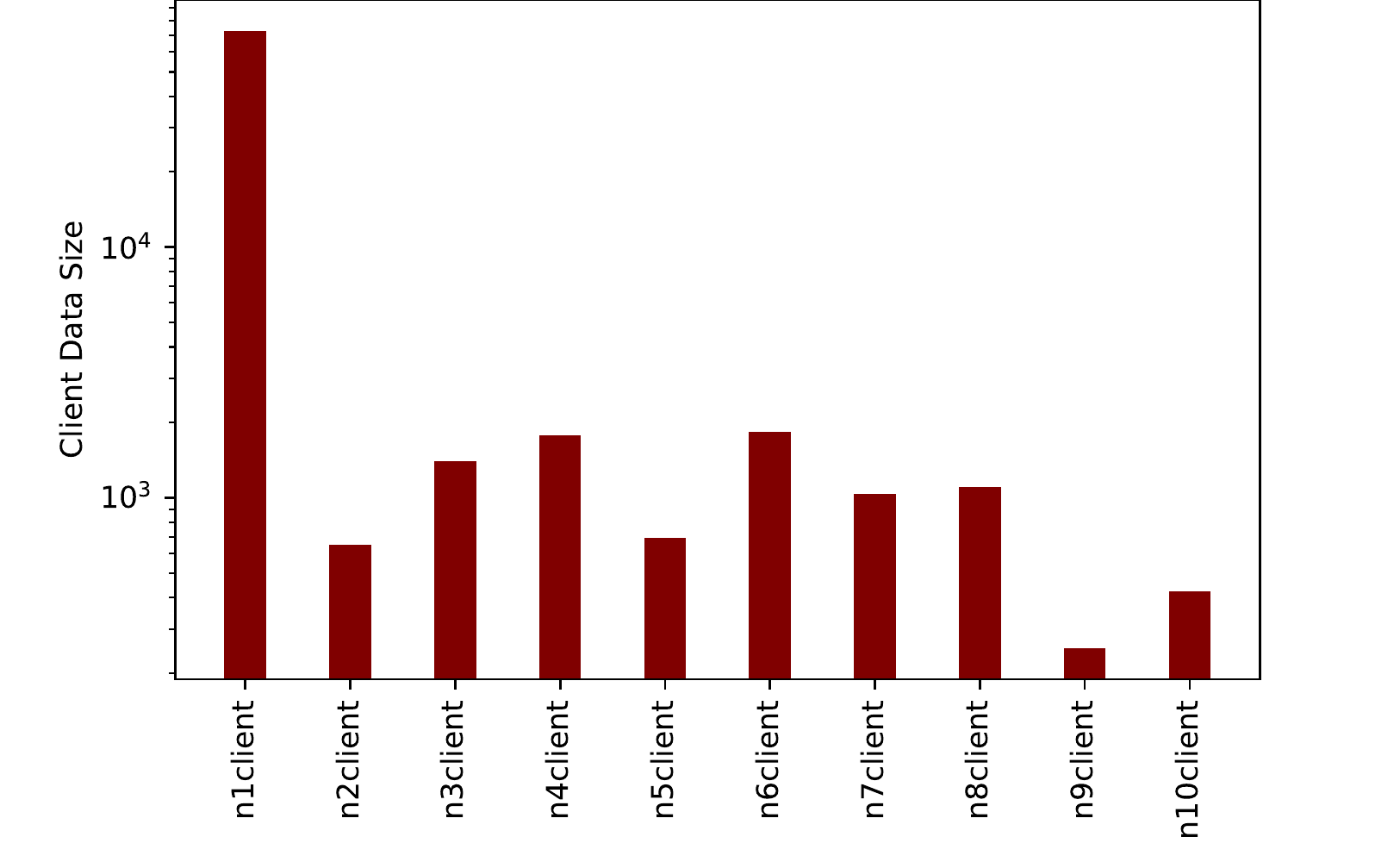}
	   \caption{The distribution of the client data.}\label{fig_datasize}
\end{figure}

\begin{figure}[htbp]
\centering
		\subfloat[KL-div distance heatmap for the CMDA method]{\includegraphics[width=100mm,scale=1.0]{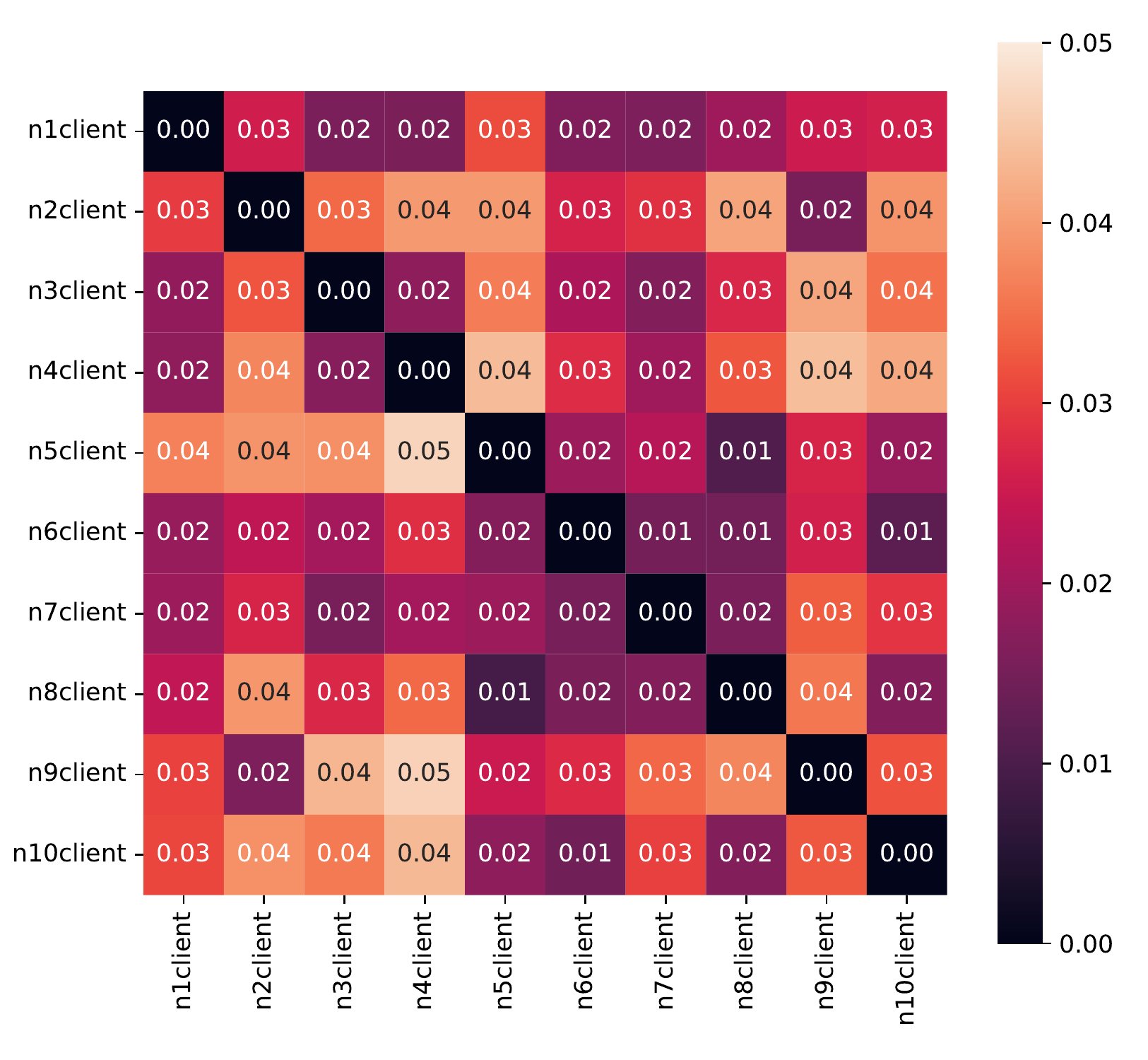}}
		\\
	    \subfloat[KL-div distance heatmap for the random splitting method]{\includegraphics[width=100mm,scale=1.0]{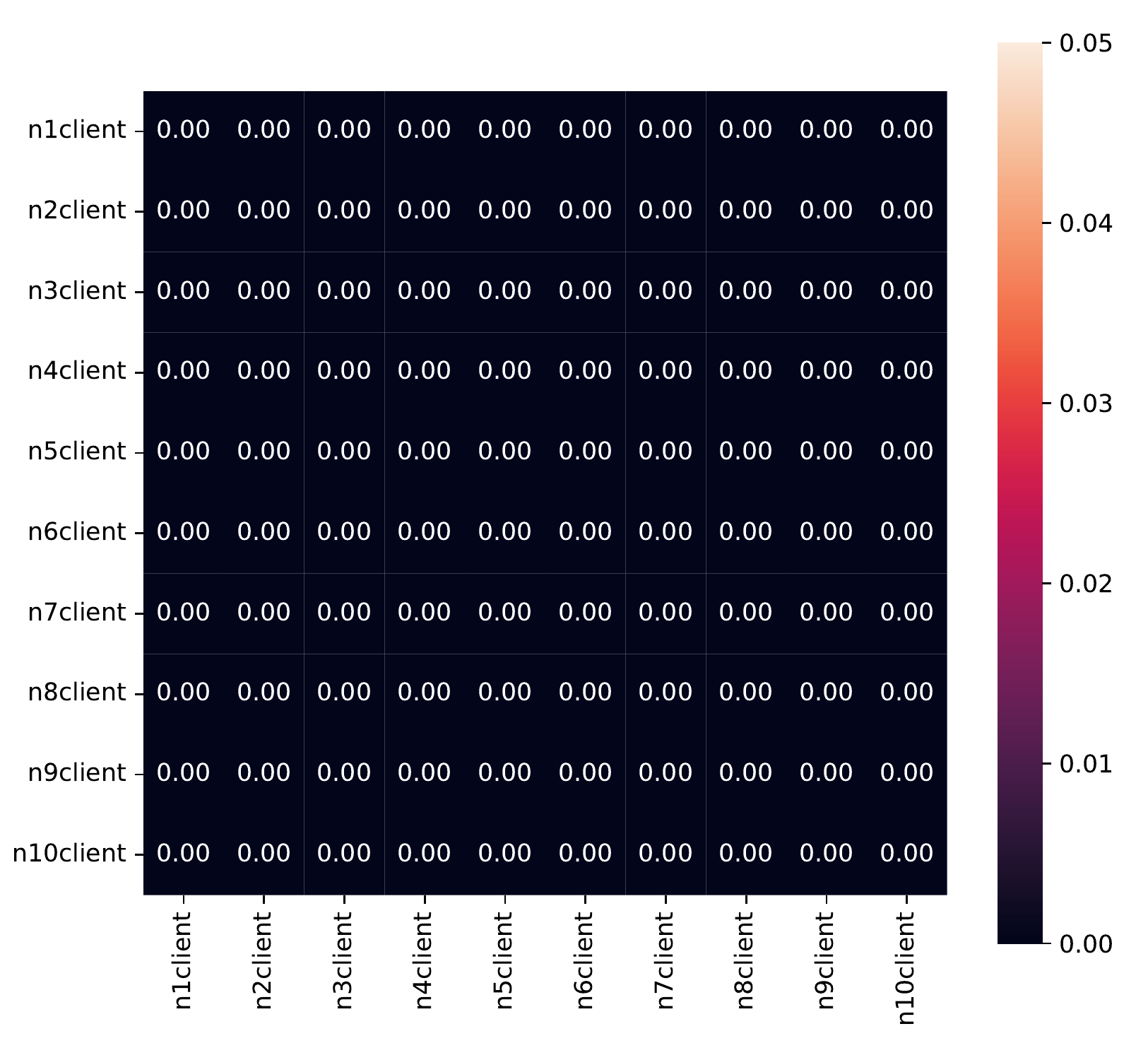}}
	  \caption{The Kullback–Leibler divergence (KL-div) distance heatmap of (a) CMDA and (b) random splitting method. 
	  The colors and values in the heatmap show the KL-div distance between two clients.
	  The heatmap (b) shows all distances are nearly zero.}\label{fig_heatmap}
\end{figure}

To evaluate the client heterogeneity generated by our proposed CMDA method. We compared CMDA with the random splitting method, which randomly assigned data samples to each client and was always used in previous studies.
The data size skew can be measured by plotting the data size of each client, as shown in Figure \ref{fig_datasize}. Client n1 contains more than half of the samples among all samples in the MS-COCO dataset.
In contrast, client n9 contains the least samples among all clients, which contain far fewer samples than other clients.
The data size difference between clients demonstrates that our clustering-based client simulation method can simulate data size skew in a federated environment.
By comparing the label distribution (Ldist) difference, we evaluate the label distribution skew of our Clustering-based Client Simulation method.
For each client, we first count the positive label occurrence of each label.
Next, we normalize the positive occurrence to [0,1] by dividing the total positive label counts and getting the normalized positive label distribution of the client. The process can be formulated as:
$$
Ldist = \frac{\sum_{n=1}^{N}y_i^l}{\sum_{n=1}^{N}\sum_{l=1}^{N_l}y_i^l}
$$
where $y_i^l$ is the binary value of $i'th$ sample at label $l$, $N$ is the total data size of client's dataset,
Moreover, $N_l$ is the total number of classes.
Last, we use Kullback–Leibler divergence to calculate the label distribution difference between clients, the heatmap of the client's label distribution difference is shown in the figure. The total clients' Kullback–Leibler divergence of k-modes and random splitting are 0.024 and 3.8e-06, respectively.
Figure \ref{fig_heatmap} demonstrates that our method creates label distribution skew between clients in the multi-label dataset compared with the random splitting method for each client. Clients n2 and n9 show the largest Kullback–Leibler divergence distance compared with clients in the k-modes client simulation method. On the other hand, clients n1 and n6 show the smallest distance compared with other clients. We will have detailed studies of these clients in the external experiment to evaluate the highest or the lowest heterogeneous clients' performance of each federated learning method.

\section{Experimental Results}\label{sec5}
In experiments, the batch size is 128, the learning rate to 1e-4, weight decay to 1e-4, the client's communication round is 4 epochs, and training for 40 epochs for all experiments.
For Per-FedAvg and pFedHN we set the meta-learning learning rate ($\beta$ in Per-FedAvg or $\alpha$ in pFedHN) equal to the learning rate. For Personal BN we personalize entire BN layers in our experiment. For KT-pFL, we set Distillation steps (R) to 1, Regularization parameter p to 0.5, and used the validation set as a public dataset. For PFADET, we set performance threshold $\lambda_1$ and $\lambda_2$ as 0.7 and 0.9.

We evaluate baseline, SOTA, and our proposed method under k-modes clustering. We first evaluated the performance of the client models and the global model. Then, we analyzed the convergent rounds and the necessary number of epochs for each method. After that, we studied the performance of the highest and the lowest label heterogeneity clients to measure how the client heterogeneity affected the federated methods.
In the experiment of hyperparameter tuning, we evaluate our under different hyperparameters $\alpha$ to find the best 
hyperparameter for FLAG.

\subsection{Overall Performance}
The evaluation metrics for multi-label classification include an average of all clients' mAP (AmAP), the worst client's mAP (WmAP), and an average global model mAP (GmAP) on clients' validation set.
As shown in Table \ref{tbl1}, the upper-bound baseline TRresNet(G) achieved the highest GmAP score 60.6\%, which are much lower than the results from the original ASL study because the data are divided into ten validation set for clients and evaluated by averaging clients' AmAP scores.
Without federated learning, the AmAP score is only around 35\%, nearly 60\% of the centralized training performance.
As shown in the tables, we can find out that pFedHN has the worst results and is even worse than local training. The reason is that our backbone model TRresNet is more complicated than a simple CNN network architecture, so the hyper network cannot learn the suitable initialized parameters. The personalized-layer-based methods PersonalBN and PersonalClassifier have similar or slightly lower performance than the FedAvg baseline.
PFADET's AmAP is similar to local training, demonstrating that their Deputy-Enhanced Transfer cannot transfer federated knowledge efficiently, and their GmAP is lower than FedAvg, showing that their deputy model cannot share federated knowledge as well.
KT-pFL provides better performance than PFADET due to the introduction of global data but cannot compete with FedAvg-based methods.

Our proposed FLAG performed the best among all methods in multi-label federated learning settings. Furthermore, we combine FLAG and the data augmentation method \cite{fedmix} for enhancement (FLAG-Aug). Compared to FedAvg, which achieved the best results among the SOTA methods, the relative improvement of FLAG-Aug on AmAP and G-mAP is 6.3\% and 8.9\%, respectively; the relative growth of FLAG on AmAP and GmAP is 4.8\% and 8.3\%, respectively.

\begin{table}[htbp]
\caption{A comparison of our proposed method and SOTA methods. The evaluation metrics are the average of clients' mAP (AmAP), the worst client's mAP (WmAP), and the average of global model mAP (GmAP) evaluated on client data. "$\pm$" after the value represents the standard deviation of clients, and "$-$" represents this method that cannot be evaluated by the metrics. We divide all methods into three groups, including non-federated baselines, TRresNet (G) and TRresNet (L), SOTA federated methods, and our proposed methods, FLAG and FLAG-Aug.}\label{tbl1}
\begin{tabular}{llll}
\toprule
\textbf{Methods} & \textbf{\textit{$AmAP (\%)$}}& \textbf{\textit{$WmAP (\%)$}}& \textbf{\textit{$GmAP (\%)$}} \\ % Table header row
\midrule
TRresNet (G) & - & $43.4$ & $60.6\pm10.3$ \\
TRresNet (L) & $33.8\pm15.7$ & $21.9$ & - \\
\midrule
FedAvg & $47.9\pm12.5$ & $31.6$ & $50.3\pm11.4$ \\
Per-FedAvg & $44.2\pm13.1$ & $27.3$ & $46.5\pm11.6$ \\
pFedHN & $13.7\pm4.8$ & $8.4$ & - \\
Personal BN & $47.3\pm12.9$ & $30.0$ & $40.2\pm16.1$ \\
Personal classifier & $45.7\pm13.1$ & $29.0$ & $49.3\pm11.2$ \\
PFADET & $33.9\pm16.1$ & $22.3$ & $21.2\pm19.1$ \\
KT-pFL & $39.0\pm13.1$ & $27.3$ & - \\
\midrule
FLAG (Ours) & $50.2\pm12.3$ & $31.2$ & $54.5\pm10.7$ \\
FLAG-Aug (Ours)  & $50.9\pm11.9$ & $29.2$ & $54.8\pm10.7$ \\
\bottomrule
\end{tabular}
\end{table}

\subsection{Early Convergence}
Convergent speed is an essential issue in federated learning. If the training loss of a model cannot be efficiently convergent, it will not be applicable to real-world applications. To measure the convergence speed of each federated learning method, we count the epochs and communication rounds that need to reach target performance.
If some federated learning methods cannot train models well to reach the target performance, we will record the epochs and communication rounds when they achieve their best performance.
We set a target performance as 80\% of the centralized learning performance, i.e., the absolute rate of target performance is 48\% of the mAP score.

\begin{figure}[htbp]
	\centering
		\subfloat[The convergent status of client models. ]{\includegraphics[width=120mm,scale=1.0]{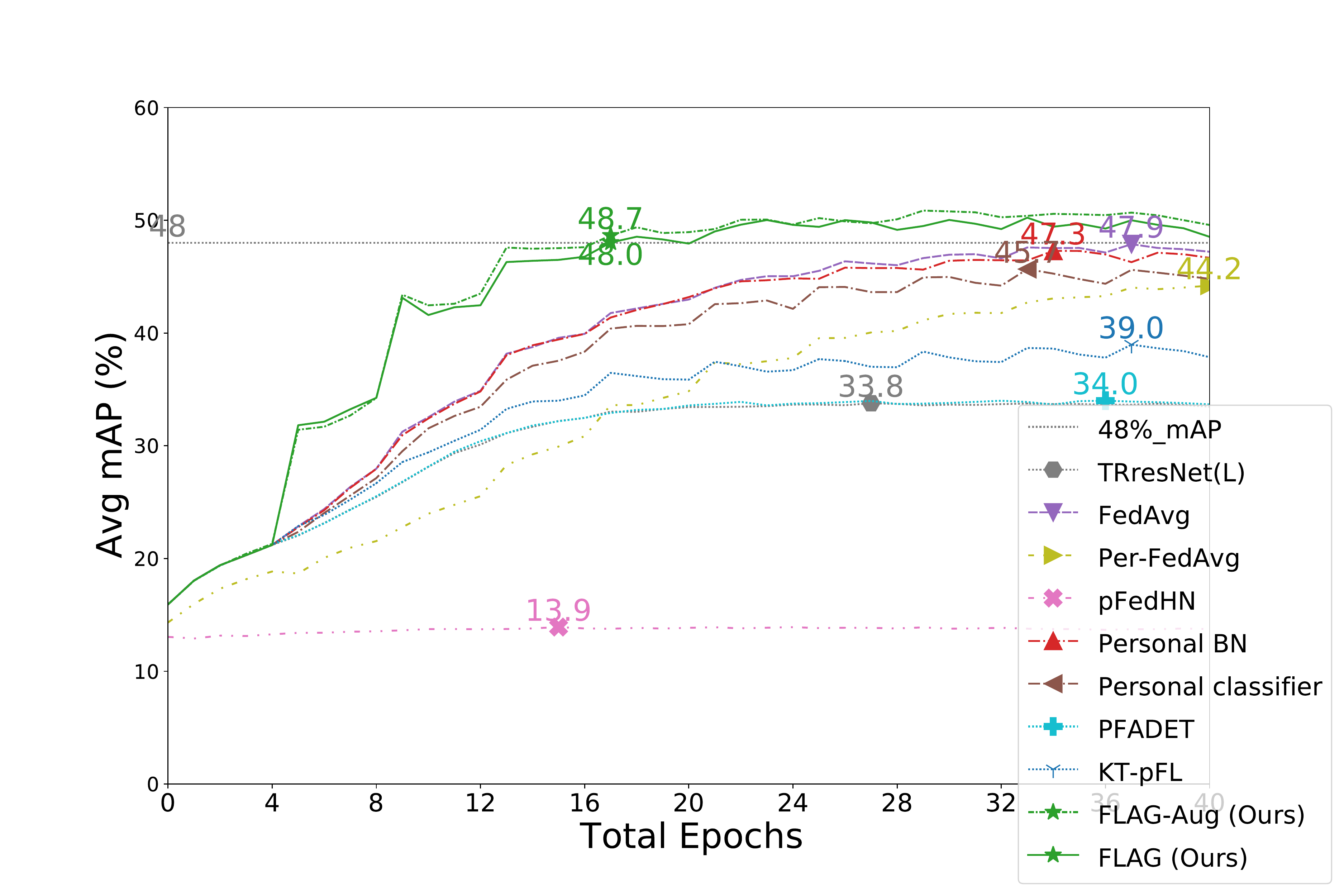}}
		\\
	    \subfloat[The convergent status of global models. ]{\includegraphics[width=120mm,scale=1.0]{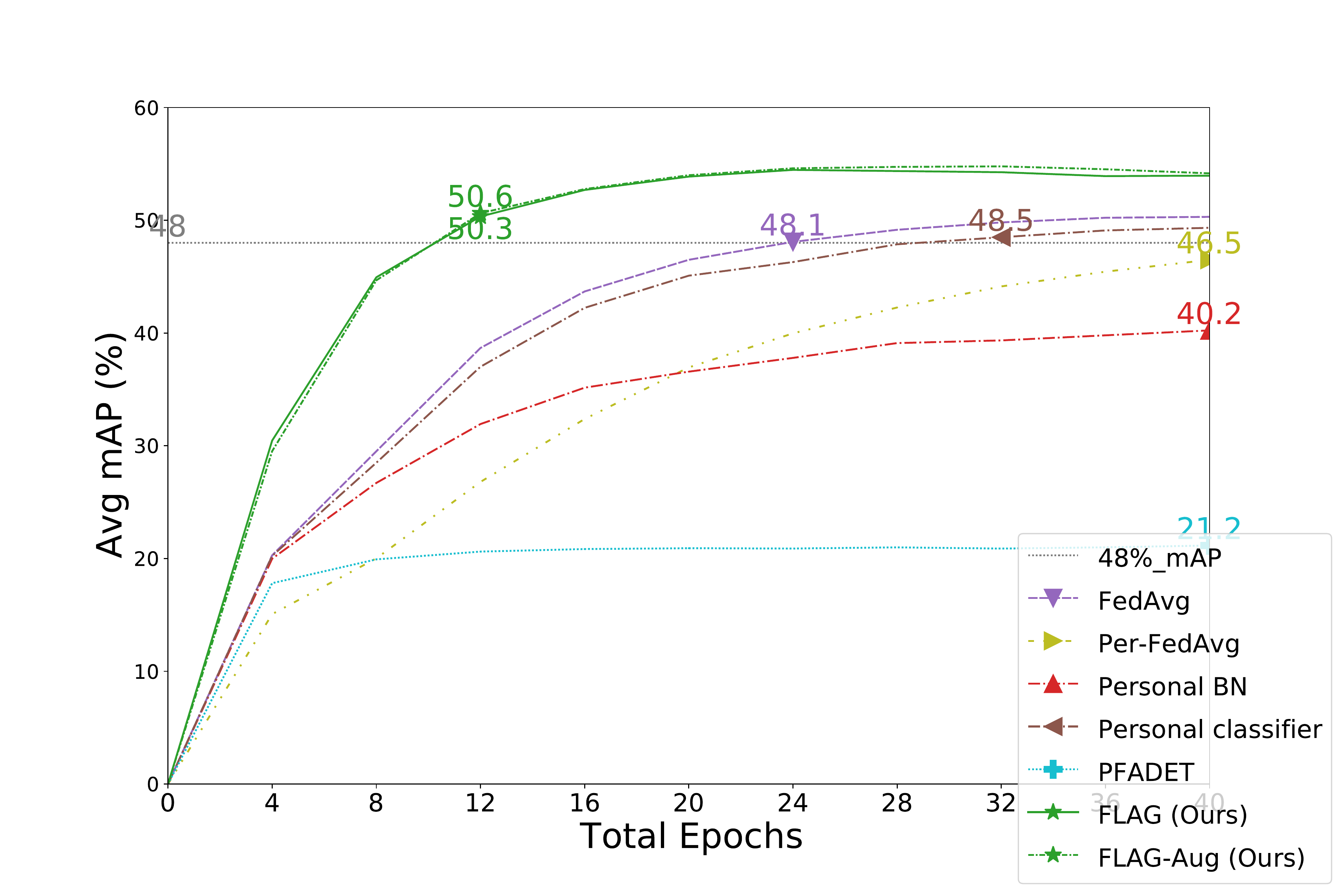}}
	  \caption{The convergent status of the client models (a) and the global model (b).
	  The marker showcase when the models reaches 48.0\% of mAP (target performance) or their best values. The gray dotted line represents the 48\% benchmark.}\label{fig_converge}
\end{figure}

As shown in Figure \ref{fig_converge}, it illustrates that our method can vastly increase the convergent speed of federated learning, which is up to two times faster than other FedAvg-based aggregation methods.
The reason for this superior convergent speed is that our method can estimate the information of each client's model through statistics of label distribution and weight them to create a better global model for the next communication round. Personalized-layer-based methods can slightly increase the converging speed. Also, we can find out that the FedAvg-based aggregation method converges faster than other aggregation methods like knowledge transfer-based or local training methods.
We conclude the reason is that directly aggregating clients' weight makes clients converge faster than other approaches. Among all FedAvg-based aggregation methods, per-FedAvg has the slowest converge speed because their meta-learning step is twice slower as other methods.
In Section \ref{client_analysis}, we look closely at different clients to measure the benefit of applying the federated learning method for a specific client.

%worst client2, client9 ; best client1, client6
\subsection{Analysis of Client Node Performance}\label{client_analysis}
\begin{figure}[htbp]
	\centering
		\subfloat[The relation between mAP and total epochs of n2 ]{\includegraphics[width=120mm,scale=1.0]{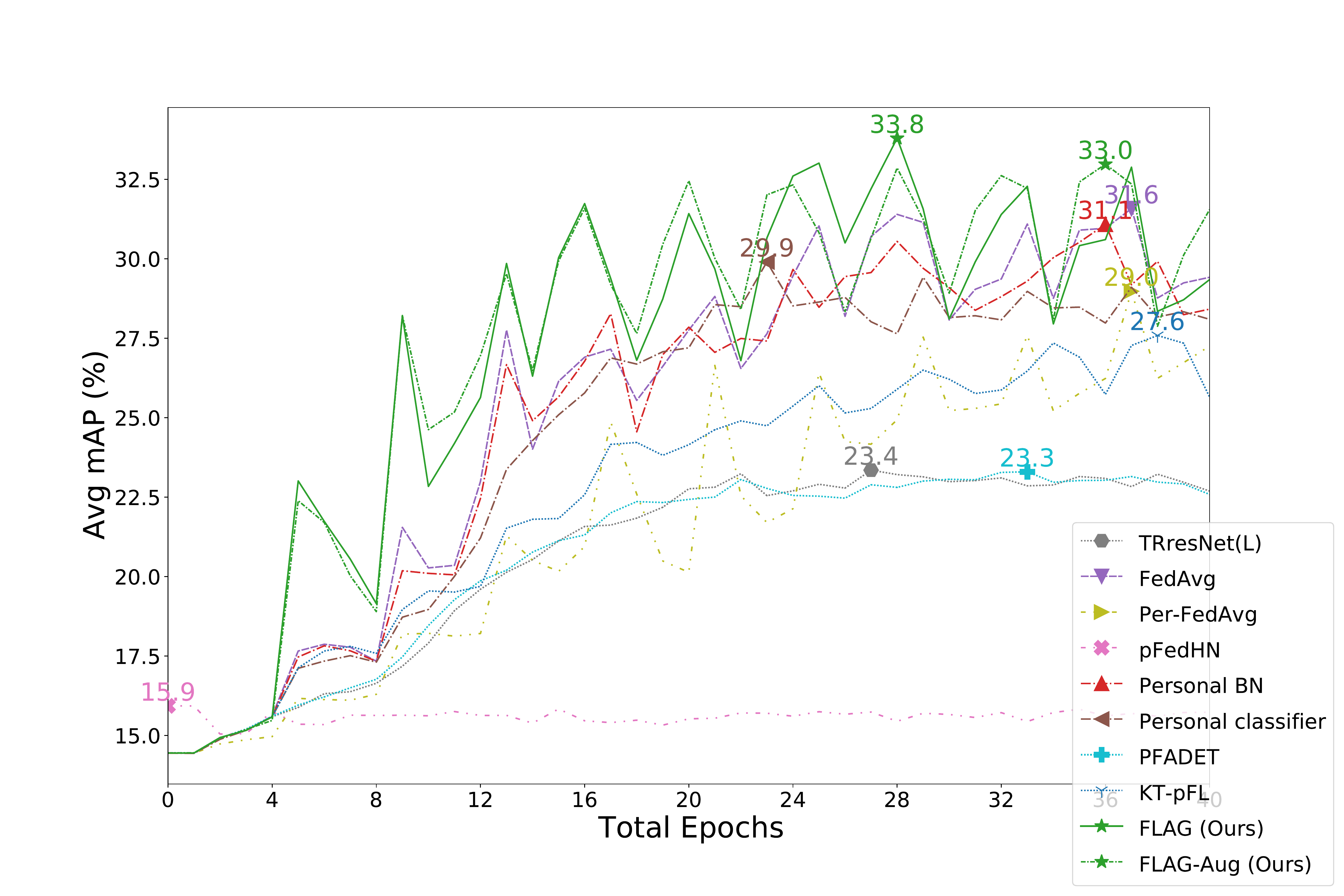}}
		\\
		\subfloat[The relation between mAP and total epochs of n9 ]{\includegraphics[width=120mm,scale=1.0]{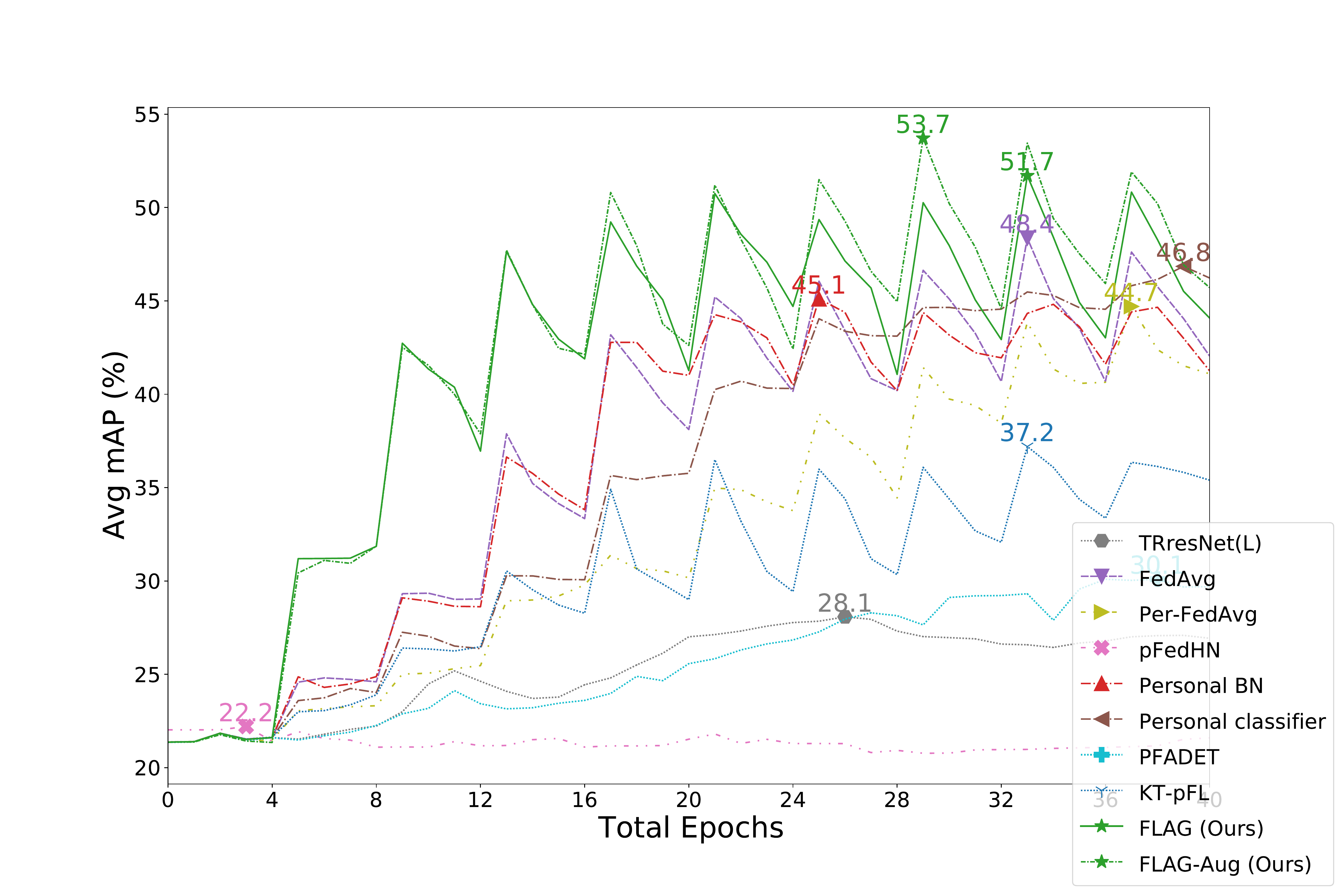}}
	  \caption{The relation between total epochs and mAP of the highest label heterogeneity clients n2 (a) and n9 (b).}\label{fig_worst_client}
\end{figure}

This section concentrates on analyzing the advantage of each federated learning method for different clients. We demonstrate the convergent speed and mAP of the highest and the lowest label heterogeneity clients from the previous section.
Figure \ref{fig_worst_client} presents that when client label heterogeneity becomes larger, during the local training process, all federated methods suffer from overfitting problems except the knowledge distillation based-methods. However, due to the federated knowledge aggregated from other clients, these methods can reach more effective models and overcome overfitting problems. In large label heterogeneity clients, our proposed FLAG methods can converge nearly twice faster as other FedAvg-based methods because FLAG methods can receive more multi-label knowledge from the clients, which has more label occurrence. Furthermore, our method can utilize multi-label knowledge from other clients to provide superior performance in clients with both small data sizes and more extensive label heterogeneity. 

\begin{figure}[htbp]
	\centering
		\subfloat[The relation between mAP and total epochs of n1
		]{\includegraphics[width=120mm,scale=1.0]{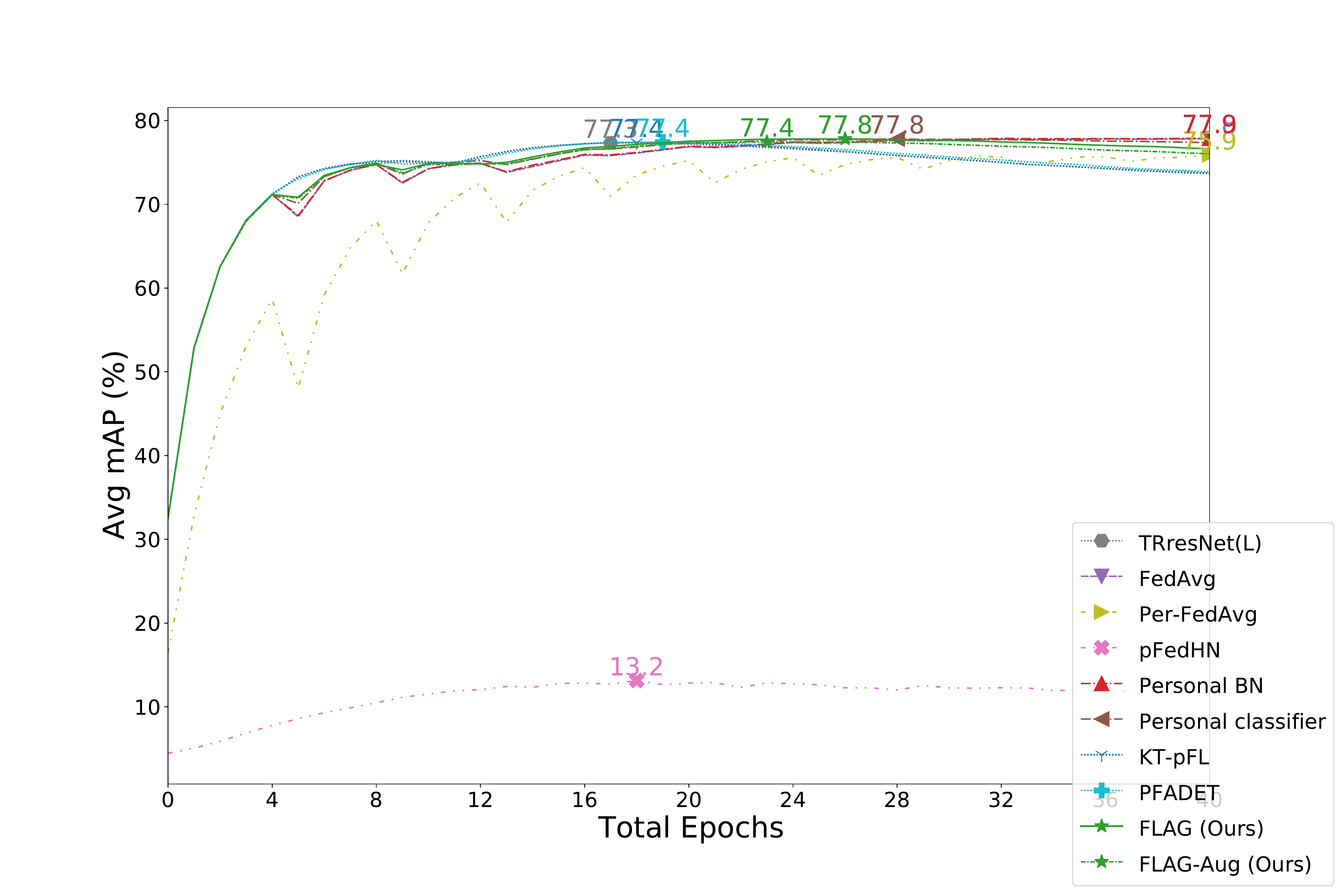}}
		\\
		\subfloat[The relation between mAP and total epochs of n6 ]{\includegraphics[width=120mm,scale=1.0]{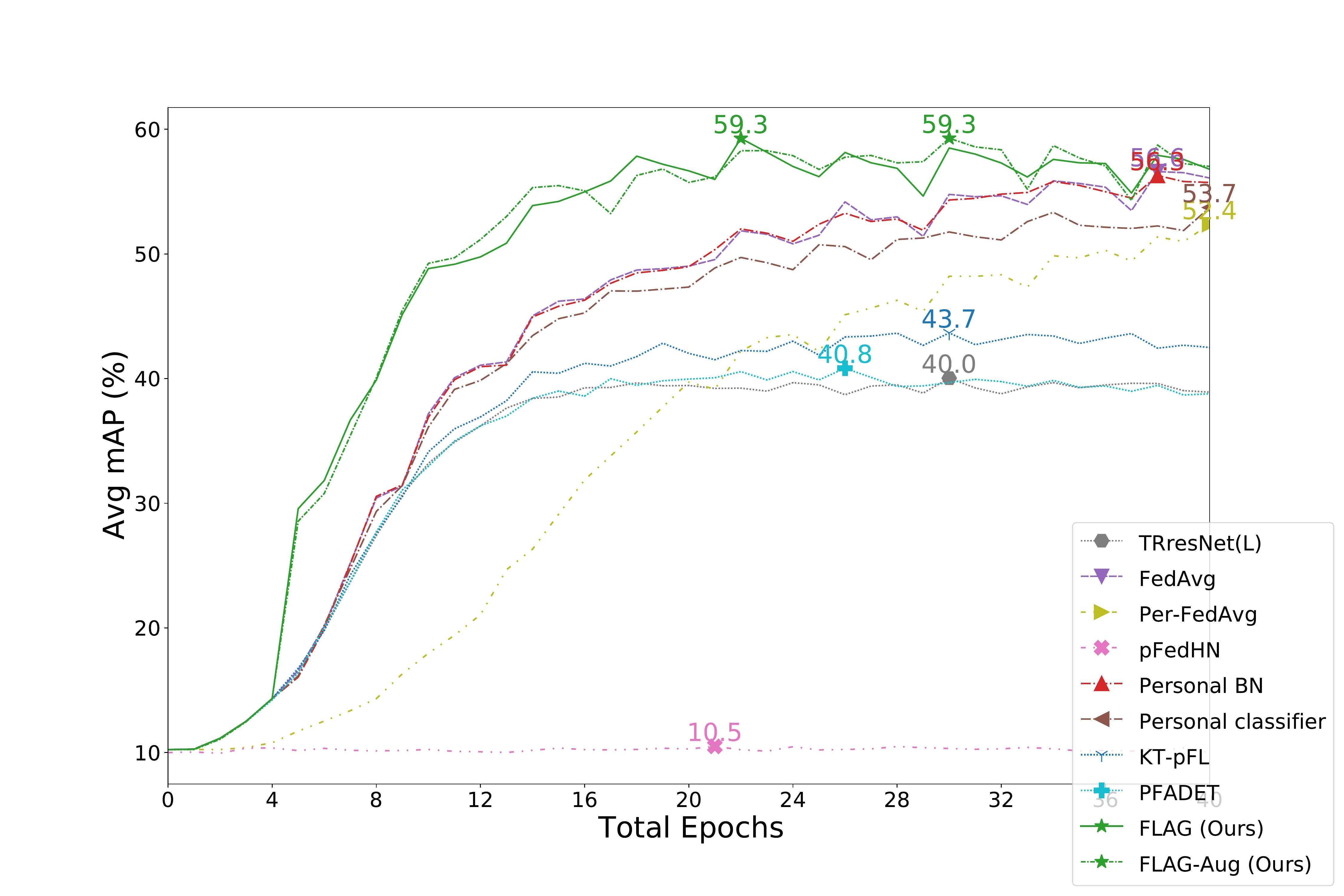}}
	  \caption{The relation between total epochs and mAP of the lowest label heterogeneity clients n1 (a) and n6 (b).}\label{fig_best_client}
\end{figure}

In Figure \ref{fig_best_client}, client n1 shows all methods receive similar final results, including local TRresNet (L); we conclude that client n1 can train well in local clients without federated methods because it has the largest data of all clients. Our method maintains a faster convergent speed for client 6, which belongs to those with less label heterogeneity, and achieves superior client mAP than the other methods.
This experiment demonstrates that our method can improve the convergent speed and classification performance of the
most- and the least-label heterogeneity clients. Especially for clients with a small dataset, our method can provide about 5\% absolute mAP and twice converge speed improvement compared with other SOTA federated learning methods.

\subsection{Hyperparameters Tuning}

In the proposed method, the hyper-parameter $\alpha$ controls the weights between label occurrence and distribution. To obtain the best hyperparameter values, we conducted a series of experiments for the FLAG method with $\alpha=[0.0,1.0]$ and evaluated their results (Table \ref{tbl2}). The GmAP scores of FLAG keep stable when $\alpha = 0.1~0.8$, and reach the top performance at  $\alpha = 0.3$.
Besides, all results in the experiments have superior performance compared to the baseline and SOTA methods.
The results demonstrate that our method can achieve high performance without detailed hyperparameter tuning.
For a balance and stable performance, we set  $\alpha = 0.3$ to compare with other SOTA methods in the experiments, as mentioned earlier.

\begin{table}[htbp]
\caption{The internal experiment result with different $\alpha$.
In this experiment, we change the value of our label distribution weighted method hyper-parameter $\alpha$.
Then evaluate the average of clients' mAP test results (AmAP) and the average of the global model evaluate on client test data result (GmAP)
with different $\alpha$ value.
}\label{tbl2}
\begin{tabular}{lll}
\toprule
\textbf{$\alpha$} & \textbf{\textit{AmAP (\%)}} & \textbf{\textit{GmAP(\%)}} \\ % Table header row
\midrule
0.0 & 48.8 & 52.2 \\
0.1 & 48.8 & 54.0 \\
0.2 & 50.8 & 54.4 \\
0.3 & 50.2 & 54.5 \\
0.4 & 49.8 & 54.9 \\
0.5 & 49.4 & 54.4 \\
0.6 & 49.0 & 54.2 \\
0.7 & 49.2 & 54.6 \\
0.8 & 48.7 & 54.7 \\
0.9 & 48.6 & 53.6 \\
1.0 & 48.4 & 53.6 \\
\bottomrule
\end{tabular}
\end{table}

\section{Discussion and Conclusion}\label{sec6}
This study proposed a new multi-label federated learning framework with a Clustering-based Multi-label Data Allocation (CMDA) method and a novel aggregation method, Fast Label-Adaptive Aggregation (FLAG),  for multi-label classification in the federated learning environment. The CMDA process can successfully simulate client heterogeneity with multi-label data, and the FLAG method outperformed the state-of-the-art methods on mAP. Moreover, FLAG can reduce training epochs (communication rounds) to achieve the best performance earlier than other methods. 

CMDA utilized correlations between labels and divided different label distributions from centralized data to simulate client heterogeneity. Kullback–Leibler divergence heatmaps of clients showed that CMDA successfully allocated multi-label data to each client, better than the conventional random splitting method to simulate the multi-label federated learning environment. 
FLAG considers label distribution and the occurrence of each client by statistics ways to take advantage of label information without data leakage problems. The experimental results demonstrated that FLAG only needs less than 50\% of training epochs and communication rounds to reach greater than or equal to the mAP of SOTA methods. These properties demonstrate a massive advantage in convergent speed. Therefore, FLAG can construct robust multi-label classification models, including global and client models.

Furthermore, when client heterogeneity increases, FLAG can maintain superior performance and provide vast converge speed improvement
compared with other methods on the highest heterogeneous client.
In the client model performance analysis, FLAG improved multi-label classification performance and the convergent speed of federated learning in different client nodes. The results also revealed that our aggregation method can still provide a helpful global model for clients, which is in heavily overfitting condition.

Even though our work has contributed much to multi-label federated learning research, some limitations can be explored to solve in the future.
The clustering method has certain randomness for simulation. Controlling the degree of client heterogeneity in federated learning is worthwhile for analysis. Currently, we use a statistical approach to obtain multi-label information from clients. However, it is a static method regarding weighting the label correlation. Thus, A dynamic weighted method during the training process is a valuable direction to make the multi-label federated learning close to real-world applications.

\section*{Acknowledgment}

We thank to National Center for High-performance Computing (NCHC) for providing computational and storage resources.

%Bibliography
\bibliographystyle{unsrt}  
\bibliography{references}  

\end{document}